\documentclass[10pt, journal]{ieeeconf}
\IEEEoverridecommandlockouts
\makeatletter 
\let\NAT@parse\undefined
\makeatother
\IEEEoverridecommandlockouts 

\usepackage{cite}
\usepackage{amsmath,amssymb,amsfonts}
\usepackage{algorithmic}
\usepackage{graphicx}
\usepackage{textcomp}
\usepackage{xcolor}
\usepackage[plainpages=false,hypertexnames=true,pdfnewwindow=true,colorlinks=true,citecolor=blue,linkcolor=black,urlcolor=blue,filecolor=blue]{hyperref}
\def\BibTeX{{\rm B\kern-.05em{\sc i\kern-.025em b}\kern-.08em
    T\kern-.1667em\lower.7ex\hbox{E}\kern-.125emX}}
\begin{document}

\title{\LARGE \bf Robotic Perception in Agri-food Manipulation: A Review}

\author{%
Jack Foster$^{1}$, Mazvydas Gudelis$^{2}$ Amir Ghalamzan E.$^{3}$\\
\textit{School of Computer Science,\\University of Lincoln, UK\\}
\thanks{\\
$^{1}$orcid.org/0000-0001-7718-4985\\
$^{2}$orcid.org/0000-0003-3101-1960\\
$^{3}$orcid.org/0000-0003-4589-0185}}

\maketitle
\begin{abstract}
To better optimise the global food supply chain, robotic solutions are needed to automate tasks currently completed by humans. Namely, phenotyping, quality analysis and harvesting are all open problems in the field of agricultural robotics. Robotic perception is a key challenge for autonomous solutions to such problems as scene understanding and object detection are vital prerequisites to any grasping tasks that a robot may undertake. This work conducts a brief review of modern robot perception models and discusses their efficacy within the agri-food domain.
\end{abstract}

\section{Introduction}
The global food chain is under increasing stress due to pressures from a growing population, climate change, and socio-political factors \cite{DBLP:journals/corr/abs-1806-06762}. Autonomous robotics are a key component of global food chain transformation. An important aspect of agri-food robotics is the application of dexterous manipulators to manage high value crop, reducing the need for skilled human workers. Robust computer vision pipelines are necessary to allow an agent to understand its environment, to identify and locate objects of interest and to complete the task it has been assigned.\\

Some of the primary perception challenges are image classification and segmentation, as well as object detection and tracking. Image classification is simply the process of assigning a class label to an image (e.g. whether or not the image contains a banana or a strawberry) \cite{lecun2015deep}. Object detection is distinct from classification as, while each object may still be classified, its location within the image is also identified and given as output from the model \cite{szegedy2013deep}. Object segmentation is similar to detection but creates a pixel-wise mask for the object instead of a bounding box representation. Object segmentation is traditionally divided into two sections: semantic and instance \cite{9356353}. The aim of semantic segmentation is to anticipate labels for object classes like strawberry and leaf, while instance segmentation separates instances of the same type, for example, distinguishing instances of all strawberry objects. The localization of such objects allows an autonomous robot to plan actions to manipulate the identified objects. Finally, object tracking is the problem of estimating the trajectory of an object
within a scene \cite{yilmaz2006object}.

Image processing pipelines designed for such tasks must balance the need for high accuracy models with the requirements found within the agricultural domain, such as the need to be executed in real-time, on robotic hardware in remote locations. In addition, models must also overcome issues concerning data availability and large perturbations in image features such as illumination, brightness, occlusions and object scale.

 \section{Existing Architectures}

While methods such as Hough transformations, random forests, support vector machines and Harr cascades have successfully been applied to computer vision tasks (including in the agricultural domain \cite{haarfortobacco}), advancements in deep learning combined with the rapid growth of available processing power has led to the domain being dominated by Neural Networks (NN). Specifically, Convolutional Neural Networks (CNNs) are well suited to image processing tasks \cite{krizhevsky2012imagenet}, as they leverage spatial information within the input vector. Furthermore, shared feature banks allow CNNs to be substantially deeper, without the exponential rise in connections \cite{lecun2015deep}, which facilitates the learning of more complex patterns.\\

Given few computational constraints, the ResNet \cite{DBLP:journals/corr/HeZRS15} and DenseNet \cite{DBLP:journals/corr/HuangLW16a} CNN architectures are rather ubiquitous in computer vision tasks, but recent work has proposed an upgrade to them. DSNet uses dense normalized shortcuts which mitigate the problem of ResNet's tight feature capacity \cite{zhang2021resnet}, while also retaining less parameters  than DenseNet and avoiding its GPU memory issues (see table \ref{tab:my-table}). This allows the model to achieve comparable results with fewer layers (DSNet50 can match a ResNet152).\\

\begin{table}[h]
\centering
\begin{tabular}{|c|c|c|c|}
\hline
Architecture & Top-1(\%) & memory (MB) & time (s) \\ \hline
ResNet50 & 24.01 & 3929 & 0.31 \\
ResNet152 & 22.16 & 7095 & 0.63 \\
DenseNet264 & 22.15 & 9981 & 0.60 \\ \hline
DSNet50 & 22.49 & 4777 & 0.37 \\
DS2Net50 & 22.03 & 5133 & 0.39 \\ \hline
\end{tabular}
\caption{GPU memory and training time on ImageNet\cite{zhang2021resnet}; memory indicates that per GPU and time indicates that per iteration}
\label{tab:my-table}
\end{table}

For object detection tasks, these models are often used as the backbone of R-CNN architectures, such as Faster R-CNN \cite{ren2015faster}, which facilitate the identification of multiple objects in a scene. While these models offer sophisticated ways of eliminating vanishing gradients in deep networks, the size and complexity of these architectures often prove prohibitive to on-board real-time execution for high-speed agricultural robotics.
\\
\\
The YOLO architectures are commonplace in agri-robotics due to their higher speed of execution, while maintaining comparable performance to the alternative models \cite{redmon2018yolov3, bochkovskiy2020yolov4}. For example, YOLOv3 is 100 times faster than Faster-RCNN. Furthermore, with the release of YOLOv4 \cite{bochkovskiy2020yolov4}, efficient CNNs have achieved a $\sim$1.5 times higher average precision score while maintaining the same frame output availability under comparable computational constraints.


\section{Labelled Data Availability}
With deep architectures outperforming traditional methods, the need for large, labelled datasets is increasingly important. While there is a growing number of agricultural datasets, there is a need to alleviate the reliance upon such datasets as the labelling process is long and tedious, and acquiring the data is often difficult, expensive, or not possible at all.
\\
As labelling is often considered a bottleneck in deploying models in new domains, there has been research on simplifying the labelling process through reducing labels to bounding boxes \cite{papandreou2015weakly} or image-level labels \cite{pinheiro2014recurrent}. While bounding boxes may be effective for well defined objects with clear structure (such as a brightly coloured fruit against a neutral backdrop), they are limited in their application to complex, noisy environments such as separating crop rows from weeds or grassland, whose boundaries are not well defined. Introduced in \cite{10.3389/fpls.2019.01404}, a semantic graphic is a graphical sketch whereby a complex object (e.g. a crop row) is represented in the form of a simplified figure (e.g. a straight green line), that is easier for the model to learn. The primary advantage of this approach is that it allows for the easy encoding of intuitive human knowledge. While this is effective, it is predicated on the na\"ive assumption that the necessary agricultural data is abundant. Often acquiring the data will be non-trivial and so alternate methods are necessary to alleviate the data dependence.\\

\begin{figure}[h]
\centering
        \includegraphics[width=0.70\linewidth]{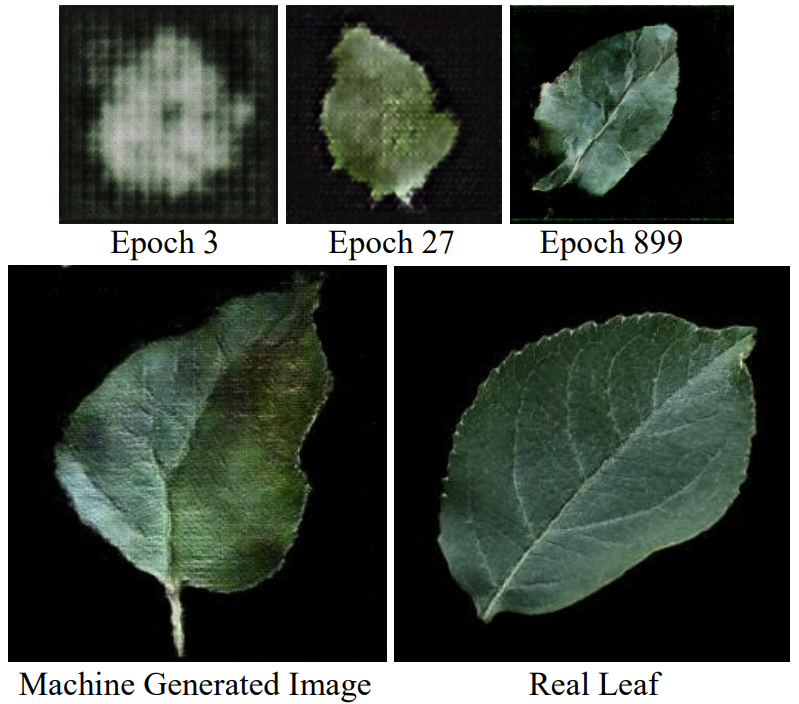}
    \caption{Fake leaf generation process \cite{radford2015unsupervised} }
    \label{156115}
\end{figure}

Another possible solution is to augment datasets through synthesising additional data-points. Since their release in 2014, Generative Adversarial Networks (GAN) \cite{goodfellow2014generative} have been seen as an appealing topic to many researchers due to their ability to be used as content creation tools. It has been widely demonstrated that GANs can be used to account for missing samples in data-sets of varying diversity. Diseases and pest infestations in agricultural crops are one of the major threats to plant quality. While the identification of these deleterious features is commonly left to machine learning models \cite{tian2019detection}, data scarcity often proves problematic when training deep networks, leading to misclassification. It is shown by the authors of \cite{radford2015unsupervised} that a Deep Convolutional Generative Adversarial Network (DCGAN) can be used to generate compellingly realistic plant leaf images (see figure \ref{156115}), which are fairly often able to deceive networks tasked with determining their legitimacy. Furthermore, an apple lesion dataset was augmented with images generated using a CycleGAN model \cite{tian2019detection}, successfully increasing the available data-pool for the detection network to train on. While the CycleGAN is capable of generating novel data, it was trained on the initial dataset and thus any generated images are likely to be perturbations of the original data. While the additional samples can improve robustness to noise in the trained classifier, it is unlikely to introduce new features that must be learned and extracted; as such, there is a requirement that the initial dataset sufficiently represents the features necessary for identifying the relevant objects. Zhang et al. \cite{zhang2021datasetgan} seek to alleviate such problems by introducing better GAN-based architectures that outperform current state-of-the-art in terms of traditional distribution quality metrics, such as the StyleGAN architecture used to generate novel images of cars, bedrooms, faces and birds \cite{karras2019style}. While not yet applied to agriculture, future work could see such models applied to crop image data generation.\\

Unsupervised pretraining may also be used, whereby a network's feature extractor layers are first trained on a large, unlabelled dataset, before their weights are frozen and the later layers can be trained on a smaller, domain-specific labelled dataset to fine-tune the network's performance. As unlabelled data does not have ground-truth labels to calculate model loss, alternative strategies must be utilised to train models.\\

Autoencoders play a fundamental role in many unsupervised learning architectures \cite{baldi2012autoencoders}. They consist of an encoder which compresses an input vector into a lower-dimensional latent representation, before a decoder attempts to reconstruct this latent representation back into the original input vector. This allows the network to be trained using the input as ground truth to the output, allowing the encoder network to learn to encode important information from the input vector, thus resulting in a feature extractor that has been pretrained on an unlabelled dataset.\\

Applications in agricultural aerial imaging have leveraged autoencoder architectures to improve robustness in iron deficiency chlorosis prediction in soybeans \cite{LI2020105557}, a problem which can lead to substantially reduced yield. Work has also been conducted in utilising convolutional autoencoders to reduce training parameters in a CNN-based plant disease detection architecture \cite{bedi2021plant}.\\
\begin{figure}[h]
\centering
        \includegraphics[width=1\linewidth]{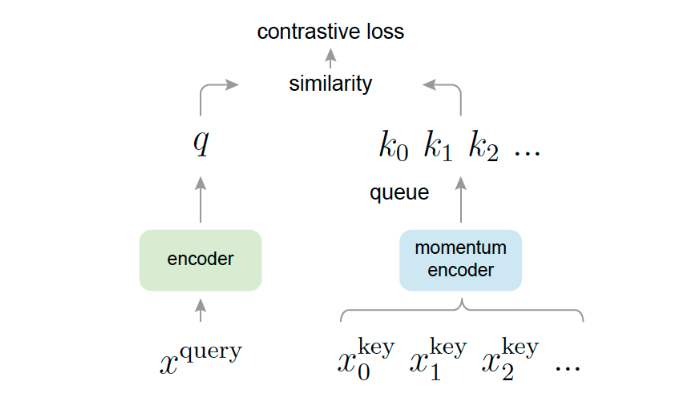}
    \caption{MoCo Architecture\cite{he2020momentum}. The left network becomes the final classifier network. The right network is the dictionary network, used to facilitate the unsupervised learning methodology.}
    \label{moco}
\end{figure}

Alternative strategies frame unsupervised learning as a dictionary look-up problem \cite{chatfield2011devil}, whereby one network encodes a stack of images into lower-dimensional latent representations (keys to the dictionary) and the primary network must then match a heavily augmented image (dictionary query) to its corresponding lower-dimensional representation. These approaches are being improved regularly, such as the use of momentum contrast learning to facilitate larger dictionary sizes while still allowing the matching of queries to up-to-date representations \cite{he2020momentum}.\\

Unsupervised training is a desirable approach because it alleviates the pressure of acquiring and labelling problem specific data, but also because it can lead to improved robustness in models \cite{LI2020105557}. This is especially true if the labelled dataset is imbalanced, as the generic feature extractor will not be conditioned on this imbalanced data. Dictionary lookup approaches have not yet seen widespread application in the agricultural domain but, like other unsupervised models, present a promising approach to developing robust object detection or classification models that can be used to aid robotic manipulation and grasping. Further work is needed to compare their efficacy against other unsupervised paradigms within the agri-foods domain.

\section{Agricultural Challenges}
Within an agricultural environment, perception can be used to solve many challenges including: phenotyping, quality analysis, and harvestability \cite{DBLP:journals/corr/abs-1806-06762}.

\subsection{Phenotyping}
Image-based plant phenotyping is concerned with the effects of the genotype and environment of a plant on its phenotype (i.e. its behaviour and characteristics) \cite{Scharr2017ICCVWorkshops}. In doing so, plant phenotyping plays a crucial role in selecting desirable traits, and designing new crops, such as ones that are resistant to environmental perturbations due to climate change \cite{10.3389/fpls.2019.01125}. The development of robust computer vision models is vital to creating a widespread, autonomous platform for plant phenotyping capable of analysing large scale plant imaging data. Common challenges within phenotyping include the counting of crop leaves \cite{buzzy2020real, prasetyo2017mango}, the building of 3D crop models \cite{Srivastava2017ICCVWorkshops}, and stem analysis \cite{Choudhury2017ICCVWorkshops}.\\

Leaf counting tasks is perhaps the most commonly addressed problem within phenotyping. Leaf counting can serve as an important indicator as to the health or growth stage of a crop, and when combined with robotic manipulation, can facilitate autonomous purging of non-desirable plants. Buzzy et al.\cite{buzzy2020real} discuss the need for on-board, real-time prediction when running on a mobile robot capable of manoeuvring through the environment autonomously. To accomplish this the YOLOv3 architecture is used, leveraging an Android phone's integrated camera as image input. Pound et al. \cite{Pound2017ICCVWorkshops} further push the link to robotic manipulation with a multi-task model that, trained on a dataset of annotated wheat images,  jointly predicts the presence of a specific phenotype, as well as the location of such features. The localisation could be leveraged to facilitate the grasping of desirable crop.\\

\begin{figure}[h]
\centering
        \includegraphics[width=0.6\linewidth]{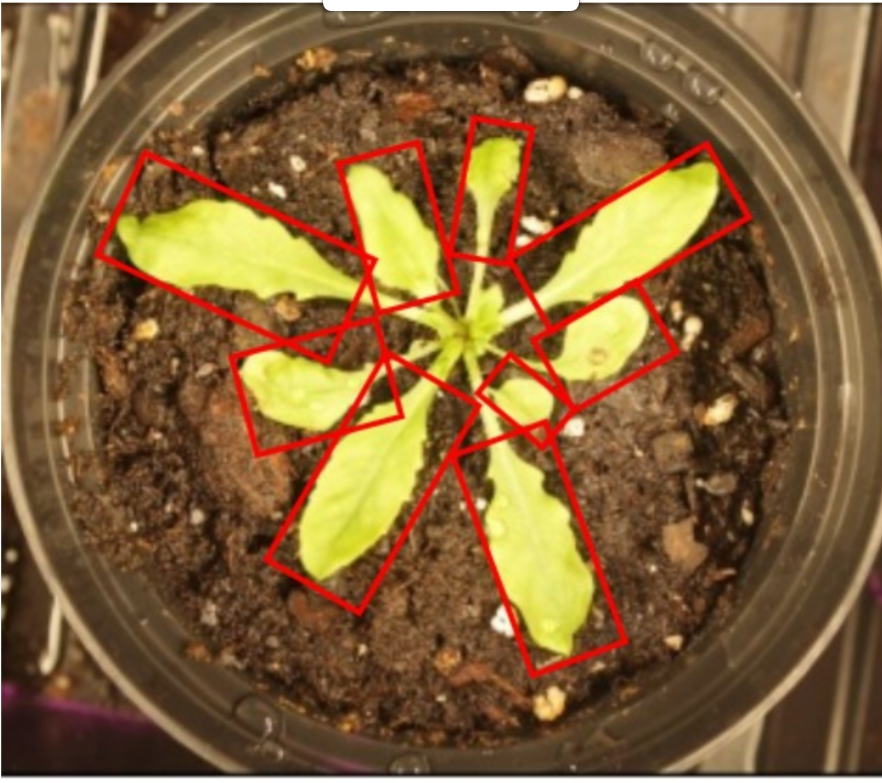}
        \caption{Leaf Detection and Bounding Box Segmentation\cite{buzzy2020real} }
        \label{leaf_count}
\end{figure}
 
3D plant models can further provide detailed measurements of plant phenomics, and can form a symbiotic relationship with robotic manipulation. As seen in the paper by Santos et al. \cite{dadalol}, dense 3D point clouds may be constructed by moving a camera around a plant, providing detailed models that can be resistant to occlusions. While \cite{dadalol} required a human operator to move the camera, it could instead be placed towards the end effector of a robot arm; allowing the robot to automate the model generation. Once the models are constructed, they may be used to improve the accuracy of grasp prediction and planning. 3D models have been used to classify draught-induced stress in plant canopies \cite{Srivastava2017ICCVWorkshops}. As plant canopys are often self-occluding and very similar, 2D models are often insufficient to accurately predict the draught-induced stress placed upon them. To achieve accurate predictions, a vision pipeline of 3D reconstruction, segmentation and feature extraction is used, leveraging deep networks at various stages of the process. While 3D models may yield better results, processing them is often more computationally expensive than their 2D counterparts, a trait that may prove problematic for real-time processing on-board high speed agricultural robotic systems. 

\subsection{Quality Analysis}
\label{QA}
The quality of a crop is of upmost importance when monitoring and maintaining produce. Quality analysis, often done by trained specialists, is the process of ensuring a crop meets a certain standard by auditing for deleterious features such as pests or disease \cite{10.3389/fpls.2020.00898}. Tedious and time consuming, automating this process has long been of interest within the agriculture sector.\\

Computer vision is a vital component to automating quality analysis, as visual aberrations are common indicators that a plant does not meet the required standards. As with other agricultural problems, deep convolutional models once again outperform alternative methods. The authors of \cite{nie2019strawberry} propose a portable hyper-spectral imaging system to detect Verticillium in strawberries, which is able to operate in real-time. In the previous case, a Faster R-CNN with a ResNet50 backbone, utilizing multi-task learning and an attention mechanism is able to detect the disease in a data-set of 3,531 images (56\% training set,
14\% validation set, and 30\% testing set) with an accuracy of 99.95\% on the 1,060 image test set. Similarly, Liu and Wang\cite{10.3389/fpls.2020.00898} detects the presence of disease and pests on tomato plants. Using a YOLOv3 architecture, augmented with multi-scale feature detection based on image pyramid, the work achieves $92.39\%$ detection accuracy with an average detection time of $20.39$ms. It should be noted, however, that the experimental setup leveraged the high speed Nvidia gtx 1080ti GPU, not typically found in lightweight robotic systems.

\subsection{Harvestability}
Utilising automation to solve harvesting tasks for high value crop is undoubtedly one of the main interests to farmers. Produce harvesting is hard to automate as it is multi-faceted, and thus well suited to human cognition. Harvesting requires produce to be identified, localised and tracked and have its harvestability predicted, before finally being grasped for harvest. If machines are employed, operators must execute harvesting commands with reference to the crop remaining, and must navigate the appliances in an optimized way to account for all produce that can be harvested. Zhang et al.\cite{ZHANG2020105384} focus on fruit tracking, exploiting a Faster R-CNN to detect apples and their branches with an accuracy of 82\%. The network also estimates shaking locations to determine which apples are certain to fall off once the tree is shook. As discussed in section \ref{QA}, Nie et al.\cite{nie2019strawberry} propose a portable hyper-spectral imaging system through the utilisation of AlexNet \cite{krizhevsky2012imagenet}, which is able to operate in real-time when classifying strawberry maturity stages and achieves a classification accuracy of 98.6\%, albeit on a rather small data-set consisting of only 240 images. Other works have integrated similar vision systems with robotic manipulators. For example, Bac et al.\cite{BAC2013148} first segment sweet-pepper plants from their background, before classifying vegetation into hard (stem, fruit) and soft (leaf, petiole) obstacles. These predictions not only allowed the sweet-pepper to be identified, but also for a collision-free path to be planned, such that a robotic manipulator may harvest the fruit.

\section{Food Industry Challenges}
In addition to the agricultural domain, robotics can also be applied to a range of problems within the food industry, classified into 3 problem domains: pick and place, packing, and serving \cite{foodreview}.

\subsection{Pick and Place}
Pick and place tasks are often integrated into the other problem domains. For example, packing requires the picking up  of an object from their starting position, and placing them into their packaging. Pick and place tasks are intertwined with the perception capabilities of the robot \cite{kasaei2018towards}, as classical solutions to pick and place tasks require either object segmentation, or object recognition and pose estimation prior to model-based grasp planning \cite{8461044}. Modern, data-driven approaches use convolutional neural networks to provide bounding boxes or segmentations, followed by pose estimation which can guide the subsequent picking up of the object \cite{7989348}, \cite{zeng2017multiview}. As a clear visual representation of the object is necessary for grasping and manipulation in other domains too, solutions may already exist to further improve the performance of food industry pick and place methods. For example, Li et al.\cite{9223541} propose an updated U-Net \cite{ronneberger2015u} architecture with a residual unit to predict the graspable parts of unknown objects by taking into account the size of the graspable part of the object and the robot's gripper, facilitating the manipulation of a wider range of objects, thus resulting in more robust grasping solutions.

\subsection{Packing}
In packing problems, robotic solutions have been largely standardised, with decision making being subject to payload specifications \cite{foodreview}. The continued development of complex pick and place grasping methodologies presents new opportunities for packing robots to undertake more nuanced challenges, such as sorting in cluttered environments where the objects are prone to occlusions; in such environments, robust perception architectures are necessary to  accurately conduct object segmentation and pose estimation. Zeng et al. \cite{zeng2017multiview} leveraged a deep CNN to estimate the 6D pose of objects present in the Amazon Picking Challenge. In this challenge, robots must pick and place objects in challenging scenarios. Primarily, the solutions must overcome: cluttered environments where there are many objects in the scene, self-occlusions due to limited camera positions, missing data due to imperfect commercial-grade sensor equipment, and the challenges around grasping small or deformable objects. This work was proposed as a solution for warehouse robotics, an area closely aligned with the packing of food goods, and thus there is scope to apply such approaches to problems within the agri-food domain.

\subsection{Serving}
Serving robotics are amongst the newest of agri-food robotics; the domain is challenging as it requires resistance to unexpected perturbations in the environment as the robot systems are human user-facing. Serving robotics are desirable as they reduce customer wait time, reduce the cost of labour, and may be more precise - reducing the number of accidents and incidents such as dropping plates or drinks \cite{malik2016review}. However, many current implementations rely upon line following techniques, paired with IR, or LIDAR to facilitate collision avoidance \cite{malik2016review, linefollowingrobot, asif2015waiter}. While these systems are fairly robust and, due to the constraints of the system, are able to operate within human environments, the development of free-moving, vision-guided robotic solutions could allow for more dynamic behaviour that allows robots to undertake more responsibility and operate with less constraints. Human aware navigation systems exist \cite{kruse2013human}, mostly leveraging robust navigation strategies such as minimum required obstacle distance or stop-and-wait strategies. While human aware navigation has seen application in domains such as assistive living \cite{vasquez2013human, rios2012intention}, they have not yet been widely applied to the food industry.
\\
\\
The development of robots for cooking is still in its infancy. This environment is challenging as it requires the high-speed and dexterous manipulation of a wide range of objects, from soft and deformable food items to knives and other dangerous cooking equipment. In other domains, robots must also operate in fast paced environments such as using conveyor belts or other sequential process dependant systems. In such cases systems tasked with object representation must provide reliable real-time data to enable the robot to physically engage with objects in its environment. For example, Kasaei et al.\cite{kasaei2016good} present a global object descriptor that allows concurrent object identification and pose estimation while balancing descriptiveness, computation time, and memory use. There is a need to extend such solutions to the domain of cooking robots to achieve deployable solutions to consumer-facing food places.

\section{Discussion}
Robotic perception strategies within the agri-foods domain have been shown to successfully solve complex challenges that would otherwise require a human worker to complete. Solving such challenges is necessary to facilitate autonomous robotic manipulation, and to further reduce the burden placed on human experts by the growing agricultural industry. There are many challenges that remain unsolved, such as access to labelled data, the trade-off between accuracy and speed of execution, and the need to further refine and optimise existing solutions to create robust, autonomous and deployable systems. While these problems remain prevalent, there are many methods and models within the field of computer vision that have yet to be applied to agricultural problems; novel solutions may be necessary for some challenges, but looking to other domains may provide answers without the need to reinvent the wheel. As technology advances, novel methods are released and often offer higher certainty, which may also reinvent the reasoning by which computer vision systems operate. It is still difficult to achieve full autonomy within the agri-food manipulation domain. Current works in perception primarily solve isolated challenges, and while these may be solved effectively, there are few solutions that prove sufficiently robust and generalisable; such that they may be deployed in real environments to alleviate the growing pressures placed upon the food supply chain.

\bibliographystyle{IEEEtran}
\bibliography{main.bbl}

\begin{thebibliography}{10}
\providecommand{\url}[1]{#1}
\csname url@samestyle\endcsname
\providecommand{\newblock}{\relax}
\providecommand{\bibinfo}[2]{#2}
\providecommand{\BIBentrySTDinterwordspacing}{\spaceskip=0pt\relax}
\providecommand{\BIBentryALTinterwordstretchfactor}{4}
\providecommand{\BIBentryALTinterwordspacing}{\spaceskip=\fontdimen2\font plus
\BIBentryALTinterwordstretchfactor\fontdimen3\font minus
  \fontdimen4\font\relax}
\providecommand{\BIBforeignlanguage}[2]{{%
\expandafter\ifx\csname l@#1\endcsname\relax
\typeout{** WARNING: IEEEtran.bst: No hyphenation pattern has been}%
\typeout{** loaded for the language `#1'. Using the pattern for}%
\typeout{** the default language instead.}%
\else
\language=\csname l@#1\endcsname
\fi
#2}}
\providecommand{\BIBdecl}{\relax}
\BIBdecl

\bibitem{DBLP:journals/corr/abs-1806-06762}
\BIBentryALTinterwordspacing
T.~Duckett, S.~Pearson, S.~Blackmore, and B.~Grieve, ``Agricultural robotics:
  The future of robotic agriculture,'' \emph{CoRR}, vol. abs/1806.06762, 2018.
  [Online]. Available: \url{http://arxiv.org/abs/1806.06762}
\BIBentrySTDinterwordspacing

\bibitem{lecun2015deep}
Y.~LeCun, Y.~Bengio, and G.~Hinton, ``Deep learning,'' \emph{nature}, vol. 521,
  no. 7553, pp. 436--444, 2015.

\bibitem{szegedy2013deep}
C.~Szegedy, A.~Toshev, and D.~Erhan, ``Deep neural networks for object
  detection,'' 2013.

\bibitem{9356353}
S.~Minaee, Y.~Y. Boykov, F.~Porikli, A.~J. Plaza, N.~Kehtarnavaz, and
  D.~Terzopoulos, ``Image segmentation using deep learning: A survey,''
  \emph{IEEE Transactions on Pattern Analysis and Machine Intelligence}, pp.
  1--1, 2021.

\bibitem{yilmaz2006object}
A.~Yilmaz, O.~Javed, and M.~Shah, ``Object tracking: A survey,'' \emph{Acm
  computing surveys (CSUR)}, vol.~38, no.~4, pp. 13--es, 2006.

\bibitem{haarfortobacco}
C.~Marzan and N.~Marcos, ``Towards tobacco leaf detection using haar cascade
  classifier and image processing techniques,'' 10 2018, pp. 63--68.

\bibitem{krizhevsky2012imagenet}
A.~Krizhevsky, I.~Sutskever, and G.~E. Hinton, ``Imagenet classification with
  deep convolutional neural networks,'' in \emph{Advances in neural information
  processing systems}, 2012, pp. 1097--1105.

\bibitem{DBLP:journals/corr/HeZRS15}
\BIBentryALTinterwordspacing
K.~He, X.~Zhang, S.~Ren, and J.~Sun, ``Deep residual learning for image
  recognition,'' \emph{CoRR}, vol. abs/1512.03385, 2015. [Online]. Available:
  \url{http://arxiv.org/abs/1512.03385}
\BIBentrySTDinterwordspacing

\bibitem{DBLP:journals/corr/HuangLW16a}
\BIBentryALTinterwordspacing
G.~Huang and Z.~L. andF Kilian Q.~Weinberger, ``Densely connected convolutional
  networks,'' \emph{CoRR}, vol. abs/1608.06993, 2016. [Online]. Available:
  \url{http://arxiv.org/abs/1608.06993}
\BIBentrySTDinterwordspacing

\bibitem{zhang2021resnet}
C.~Zhang, P.~Benz, D.~M. Argaw, S.~Lee, J.~Kim, F.~Rameau, J.-C. Bazin, and
  I.~S. Kweon, ``Resnet or densenet? introducing dense shortcuts to resnet,''
  in \emph{Proceedings of the IEEE/CVF Winter Conference on Applications of
  Computer Vision}, 2021, pp. 3550--3559.

\bibitem{ren2015faster}
S.~Ren, K.~He, R.~Girshick, and J.~Sun, ``Faster r-cnn: Towards real-time
  object detection with region proposal networks,'' \emph{arXiv preprint
  arXiv:1506.01497}, 2015.

\bibitem{redmon2018yolov3}
J.~Redmon and A.~Farhadi, ``Yolov3: An incremental improvement,'' 2018.

\bibitem{bochkovskiy2020yolov4}
A.~Bochkovskiy, C.-Y. Wang, and H.-Y.~M. Liao, ``Yolov4: Optimal speed and
  accuracy of object detection,'' 2020.

\bibitem{papandreou2015weakly}
G.~Papandreou, L.-C. Chen, K.~P. Murphy, and A.~L. Yuille, ``Weakly-and
  semi-supervised learning of a deep convolutional network for semantic image
  segmentation,'' in \emph{Proceedings of the IEEE international conference on
  computer vision}, 2015, pp. 1742--1750.

\bibitem{pinheiro2014recurrent}
P.~Pinheiro and R.~Collobert, ``Recurrent convolutional neural networks for
  scene labeling,'' in \emph{International conference on machine
  learning}.\hskip 1em plus 0.5em minus 0.4em\relax PMLR, 2014, pp. 82--90.

\bibitem{10.3389/fpls.2019.01404}
\BIBentryALTinterwordspacing
S.~P. Adhikari, H.~Yang, and H.~Kim, ``Learning semantic graphics using
  convolutional encoder–decoder network for autonomous weeding in paddy,''
  \emph{Frontiers in Plant Science}, vol.~10, p. 1404, 2019. [Online].
  Available: \url{https://www.frontiersin.org/article/10.3389/fpls.2019.01404}
\BIBentrySTDinterwordspacing

\bibitem{goodfellow2014generative}
I.~J. Goodfellow, J.~Pouget-Abadie, M.~Mirza, B.~Xu, D.~Warde-Farley, S.~Ozair,
  A.~Courville, and Y.~Bengio, ``Generative adversarial networks,'' \emph{arXiv
  preprint arXiv:1406.2661}, 2014.

\bibitem{tian2019detection}
Y.~Tian, G.~Yang, Z.~Wang, E.~Li, and Z.~Liang, ``Detection of apple lesions in
  orchards based on deep learning methods of cyclegan and yolov3-dense,''
  \emph{Journal of Sensors}, vol. 2019, 2019.

\bibitem{radford2015unsupervised}
A.~Radford, L.~Metz, and S.~Chintala, ``Unsupervised representation learning
  with deep convolutional generative adversarial networks,'' \emph{arXiv
  preprint arXiv:1511.06434}, 2015.

\bibitem{zhang2021datasetgan}
Y.~Zhang, H.~Ling, J.~Gao, K.~Yin, J.-F. Lafleche, A.~Barriuso, A.~Torralba,
  and S.~Fidler, ``Datasetgan: Efficient labeled data factory with minimal
  human effort,'' \emph{arXiv preprint arXiv:2104.06490}, 2021.

\bibitem{karras2019style}
T.~Karras, S.~Laine, and T.~Aila, ``A style-based generator architecture for
  generative adversarial networks,'' in \emph{Proceedings of the IEEE/CVF
  Conference on Computer Vision and Pattern Recognition}, 2019, pp. 4401--4410.

\bibitem{baldi2012autoencoders}
P.~Baldi, ``Autoencoders, unsupervised learning, and deep architectures,'' in
  \emph{Proceedings of ICML workshop on unsupervised and transfer
  learning}.\hskip 1em plus 0.5em minus 0.4em\relax JMLR Workshop and
  Conference Proceedings, 2012, pp. 37--49.

\bibitem{LI2020105557}
\BIBentryALTinterwordspacing
J.~Li, C.~Oswald, G.~L. Graef, and Y.~Shi, ``Improving model robustness for
  soybean iron deficiency chlorosis rating by unsupervised pre-training on
  unmanned aircraft system derived images,'' \emph{Computers and Electronics in
  Agriculture}, vol. 175, p. 105557, 2020. [Online]. Available:
  \url{https://www.sciencedirect.com/science/article/pii/S016816992030483X}
\BIBentrySTDinterwordspacing

\bibitem{bedi2021plant}
P.~Bedi and P.~Gole, ``Plant disease detection using hybrid model based on
  convolutional autoencoder and convolutional neural network,''
  \emph{Artificial Intelligence in Agriculture}, 2021.

\bibitem{he2020momentum}
K.~He, H.~Fan, Y.~Wu, S.~Xie, and R.~Girshick, ``Momentum contrast for
  unsupervised visual representation learning,'' in \emph{Proceedings of the
  IEEE/CVF Conference on Computer Vision and Pattern Recognition}, 2020, pp.
  9729--9738.

\bibitem{chatfield2011devil}
K.~Chatfield, V.~S. Lempitsky, A.~Vedaldi, and A.~Zisserman, ``The devil is in
  the details: an evaluation of recent feature encoding methods.'' in
  \emph{BMVC}, vol.~2, no.~4, 2011, p.~8.

\bibitem{Scharr2017ICCVWorkshops}
H.~Scharr, T.~P. Pridmore, and S.~A. Tsaftaris, ``Editorial: Computer vision
  problems in plant phenotyping, cvppp 2017 -- introduction to the cvppp 2017
  workshop papers,'' in \emph{Proceedings of the IEEE International Conference
  on Computer Vision (ICCV) Workshops}, Oct 2017.

\bibitem{10.3389/fpls.2019.01125}
\BIBentryALTinterwordspacing
J.~M. Costa, J.~Marques~da Silva, C.~Pinheiro, M.~Barón, P.~Mylona,
  M.~Centritto, M.~Haworth, F.~Loreto, B.~Uzilday, I.~Turkan, and M.~M.
  Oliveira, ``Opportunities and limitations of crop phenotyping in southern
  european countries,'' \emph{Frontiers in Plant Science}, vol.~10, p. 1125,
  2019. [Online]. Available:
  \url{https://www.frontiersin.org/article/10.3389/fpls.2019.01125}
\BIBentrySTDinterwordspacing

\bibitem{buzzy2020real}
M.~Buzzy, V.~Thesma, M.~Davoodi, and J.~Mohammadpour~Velni, ``Real-time plant
  leaf counting using deep object detection networks,'' \emph{Sensors},
  vol.~20, no.~23, p. 6896, 2020.

\bibitem{prasetyo2017mango}
E.~Prasetyo, R.~D. Adityo, N.~Suciati, and C.~Fatichah, ``Mango leaf image
  segmentation on hsv and ycbcr color spaces using otsu thresholding,'' in
  \emph{2017 3rd International Conference on Science and Technology-Computer
  (ICST)}.\hskip 1em plus 0.5em minus 0.4em\relax IEEE, 2017, pp. 99--103.

\bibitem{Srivastava2017ICCVWorkshops}
S.~Srivastava, S.~Bhugra, B.~Lall, and S.~Chaudhury, ``Drought stress
  classification using 3d plant models,'' in \emph{Proceedings of the IEEE
  International Conference on Computer Vision (ICCV) Workshops}, Oct 2017.

\bibitem{Choudhury2017ICCVWorkshops}
S.~Das~Choudhury, S.~Goswami, S.~Bashyam, A.~Samal, and T.~Awada, ``Automated
  stem angle determination for temporal plant phenotyping analysis,'' in
  \emph{Proceedings of the IEEE International Conference on Computer Vision
  (ICCV) Workshops}, Oct 2017.

\bibitem{Pound2017ICCVWorkshops}
M.~P. Pound, J.~A. Atkinson, D.~M. Wells, T.~P. Pridmore, and A.~P. French,
  ``Deep learning for multi-task plant phenotyping,'' in \emph{Proceedings of
  the IEEE International Conference on Computer Vision (ICCV) Workshops}, Oct
  2017.

\bibitem{dadalol}
T.~T. Santos, L.~V. Koenigkan, J.~G.~A. Barbedo, and G.~C. Rodrigues, ``3d
  plant modeling: Localization, mapping and segmentation for plant phenotyping
  using a single hand-held camera,'' in \emph{Computer Vision - ECCV 2014
  Workshops}, L.~Agapito, M.~M. Bronstein, and C.~Rother, Eds.\hskip 1em plus
  0.5em minus 0.4em\relax Cham: Springer International Publishing, 2015, pp.
  247--263.

\bibitem{10.3389/fpls.2020.00898}
\BIBentryALTinterwordspacing
J.~Liu and X.~Wang, ``Tomato diseases and pests detection based on improved
  yolo v3 convolutional neural network,'' \emph{Frontiers in Plant Science},
  vol.~11, p. 898, 2020. [Online]. Available:
  \url{https://www.frontiersin.org/article/10.3389/fpls.2020.00898}
\BIBentrySTDinterwordspacing

\bibitem{nie2019strawberry}
X.~Nie, L.~Wang, H.~Ding, and M.~Xu, ``Strawberry verticillium wilt detection
  network based on multi-task learning and attention,'' \emph{IEEE Access},
  vol.~7, pp. 170\,003--170\,011, 2019.

\bibitem{ZHANG2020105384}
\BIBentryALTinterwordspacing
J.~Zhang, M.~Karkee, Q.~Zhang, X.~Zhang, M.~Yaqoob, L.~Fu, and S.~Wang,
  ``Multi-class object detection using faster r-cnn and estimation of shaking
  locations for automated shake-and-catch apple harvesting,'' \emph{Computers
  and Electronics in Agriculture}, vol. 173, p. 105384, 2020. [Online].
  Available:
  \url{https://www.sciencedirect.com/science/article/pii/S0168169919327073}
\BIBentrySTDinterwordspacing

\bibitem{BAC2013148}
\BIBentryALTinterwordspacing
C.~Bac, J.~Hemming, and E.~{van Henten}, ``Robust pixel-based classification of
  obstacles for robotic harvesting of sweet-pepper,'' \emph{Computers and
  Electronics in Agriculture}, vol.~96, pp. 148--162, 2013. [Online].
  Available:
  \url{https://www.sciencedirect.com/science/article/pii/S0168169913001099}
\BIBentrySTDinterwordspacing

\bibitem{foodreview}
J.~Iqbal, Z.~H. Khan, and A.~Khalid, ``Prospects of robotics in food
  industry,'' \emph{Food Science and Technology (Campinas)}, vol.~37, pp.
  159--165, 04 2017.

\bibitem{kasaei2018towards}
S.~H. Kasaei, M.~Oliveira, G.~H. Lim, L.~S. Lopes, and A.~M. Tom{\'e},
  ``Towards lifelong assistive robotics: A tight coupling between object
  perception and manipulation,'' \emph{Neurocomputing}, vol. 291, pp. 151--166,
  2018.

\bibitem{8461044}
A.~Zeng, S.~Song, K.-T. Yu, E.~Donlon, F.~R. Hogan, M.~Bauza, D.~Ma, O.~Taylor,
  M.~Liu, E.~Romo, N.~Fazeli, F.~Alet, N.~C. Dafle, R.~Holladay, I.~Morena,
  P.~Qu~Nair, D.~Green, I.~Taylor, W.~Liu, T.~Funkhouser, and A.~Rodriguez,
  ``Robotic pick-and-place of novel objects in clutter with multi-affordance
  grasping and cross-domain image matching,'' in \emph{2018 IEEE International
  Conference on Robotics and Automation (ICRA)}, 2018, pp. 3750--3757.

\bibitem{7989348}
M.~Schwarz, A.~Milan, C.~Lenz, A.~Muñoz, A.~S. Periyasamy, M.~Schreiber,
  S.~Schüller, and S.~Behnke, ``Nimbro picking: Versatile part handling for
  warehouse automation,'' in \emph{2017 IEEE International Conference on
  Robotics and Automation (ICRA)}, 2017, pp. 3032--3039.

\bibitem{zeng2017multiview}
A.~Zeng, K.-T. Yu, S.~Song, D.~Suo, E.~W.~J. au2, A.~Rodriguez, and J.~Xiao,
  ``Multi-view self-supervised deep learning for 6d pose estimation in the
  amazon picking challenge,'' 2017.

\bibitem{9223541}
Y.~Li, L.~Schomaker, and S.~H. Kasaei, ``Learning to grasp 3d objects using
  deep residual u-nets,'' in \emph{2020 29th IEEE International Conference on
  Robot and Human Interactive Communication (RO-MAN)}, 2020, pp. 781--787.

\bibitem{ronneberger2015u}
O.~Ronneberger, P.~Fischer, and T.~Brox, ``U-net: Convolutional networks for
  biomedical image segmentation,'' in \emph{International Conference on Medical
  image computing and computer-assisted intervention}.\hskip 1em plus 0.5em
  minus 0.4em\relax Springer, 2015, pp. 234--241.

\bibitem{malik2016review}
N.~Malik, N.~Rani, P.~A. Singh, and S.~Pragya, ``Review paper on-serving robot
  new generation electronic waiter,'' \emph{IJIRST--International Journal for
  Innovative Research in Science \& Technology}, vol.~2, no.~11, pp. 775--777,
  2016.

\bibitem{linefollowingrobot}
A.~Hamid, S.~Hamdany, L.~Albak, R.~Rafi, and R.~Al-Nima, ``Wireless waiter
  robot,'' 12 2019.

\bibitem{asif2015waiter}
M.~Asif, M.~Sabeel, and K.~Mujeeb-ur Rahman, ``Waiter robot-solution to
  restaurant automation,'' in \emph{Proceedings of the 1st student multi
  disciplinary research conference (MDSRC), At Wah, Pakistan}, 2015, pp.
  14--15.

\bibitem{kruse2013human}
T.~Kruse, A.~K. Pandey, R.~Alami, and A.~Kirsch, ``Human-aware robot
  navigation: A survey,'' \emph{Robotics and Autonomous Systems}, vol.~61,
  no.~12, pp. 1726--1743, 2013.

\bibitem{vasquez2013human}
D.~Vasquez, P.~Stein, J.~Rios-Martinez, A.~Escobedo, A.~Spalanzani, and
  C.~Laugier, ``Human aware navigation for assistive robotics,'' in
  \emph{Experimental Robotics}.\hskip 1em plus 0.5em minus 0.4em\relax
  Springer, 2013, pp. 449--462.

\bibitem{rios2012intention}
J.~Rios-Martinez, A.~Escobedo, A.~Spalanzani, and C.~Laugier, ``Intention
  driven human aware navigation for assisted mobility,'' in \emph{Workshop on
  Assistance and Service robotics in a human environment at IROS}, 2012.

\bibitem{kasaei2016good}
S.~H. Kasaei, A.~M. Tom{\'e}, L.~S. Lopes, and M.~Oliveira, ``Good: A global
  orthographic object descriptor for 3d object recognition and manipulation,''
  \emph{Pattern Recognition Letters}, vol.~83, pp. 312--320, 2016.

\end{thebibliography}

\end{document}